\documentclass[lettersize,journal]{IEEEtran}

\usepackage{amsmath,amsfonts}
\usepackage{algorithmic}
\usepackage{algorithm}
\usepackage{array}

\usepackage[space]{cite}
\usepackage{amssymb}
\usepackage{graphicx}
\usepackage{textcomp}
\usepackage{xcolor}
\usepackage{caption}
\usepackage{amsthm,bm} %
\usepackage{multirow}
\usepackage{array}
\usepackage{graphicx}
\usepackage{array,multirow}
\usepackage{booktabs}
\usepackage{caption}
\usepackage{subcaption}

\usepackage{stfloats}
\usepackage{url}
\usepackage{verbatim}
\usepackage{microtype}

\begin{document}

\title{Significance of Anatomical Constraints in Virtual Try-On\\
}

\author{Debapriya~Roy,
        Sanchayan~Santra,
        Diganta~Mukherjee,
        and~Bhabatosh~Chanda
\thanks{Manuscript received Month XX, XXXX; revised Month XX, XXXX.}%
\thanks{D. Roy is with the dept. of Institute of Advancing Intelligence at the TCG Centres for Research and Education in Science and Technology, Kolkata, India. (E-mail: debapriyakundu1@gmail.com)}%
\thanks{S. Santra is with the Institute for Datability Science, Osaka University, Osaka, Japan. (E-mail: sanchayan.santra@gmail.com)}%
\thanks{Prof. Mukherjee is with the Indian Statistical Institute, Kolkata, India. (E-mail: diganta@isical.ac.in)}%
\thanks{Prof. Chanda was formerly with the Indian Statistical Institute, Kolkata, India, and currently, he is with Indian Institute of Information Technology Kalyani, India. (E-mail: bchanda57@gmail.com)}
}
\markboth{IEEE TRANSACTIONS ON EMERGING TOPICS IN COMPUTATIONAL INTELLIGENCE}%
{Roy \MakeLowercase{\textit{et al.}}: Significance of Anatomical Constraints in Virtual Try-On}

\IEEEpubid{0000--0000/00\$00.00~\copyright~2023 IEEE}
\maketitle

\begin{abstract}
The system of Virtual Try-ON (VTON) allows a user to try a product virtually. In general, a VTON system takes a clothing source and a person's image to predict the try-on output of the person in the given clothing. Although existing methods perform well for simple poses, in case of bent or crossed arms posture or when there is a significant difference between the alignment of the source clothing and the pose of the target person, these methods fail by generating inaccurate clothing deformations. In the VTON methods that employ Thin Plate Spline (TPS) based clothing transformations, this mainly occurs for two reasons - (1)~the second-order smoothness constraint of TPS that restricts the bending of the object plane. (2)~Overlaps among different clothing parts (e.g., sleeves and torso) can not be modeled by a single TPS transformation, as it assumes the clothing as a single planar object; therefore, disregards the independence of movement of different clothing parts. To this end, we make two major contributions. Concerning the bending limitations of TPS, we propose a human AnaTomy-Aware Geometric (ATAG) transformation.  Regarding the overlap issue, we propose a part-based warping approach that divides the clothing into independently warpable parts to warp them separately and later combine them. Extensive analysis shows the efficacy of this approach.
\end{abstract}


\section{Introduction}
\label{intro}
\IEEEPARstart{V}irtual Try-ON (VTON) is a promising application of human-centered image synthesis. Computationally, it is a method of synthesizing an image of a query person wearing a given clothing, thus solving the issue of the trial of clothing in the case of online shopping. While it has some great benefits from an application perspective, attempting the problem is challenging in terms of maintaining the photo-realism of the output mainly in the case of complex postures and exploiting easy-accessible data. 

\begin{figure}[!t]
	\centering
	\includegraphics[width=\linewidth]{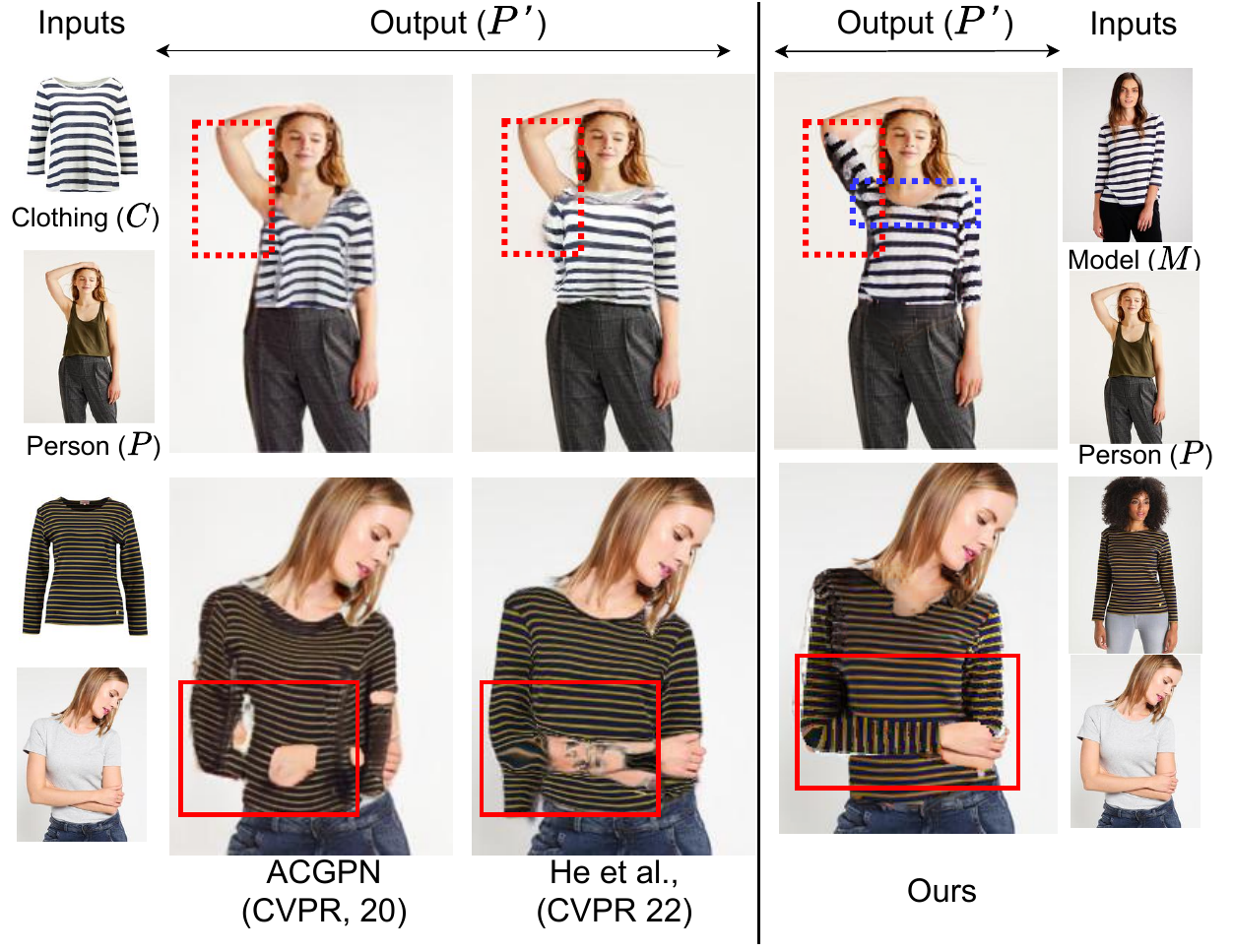}
	\captionof{figure}{
	A comparative study with some benchmark methods, ACGPN (Warping-based, C2P) and He et al. (Flow-Based, C2P), illustrating their lack of ability in warping the sleeves in case of significant arm bending (first row), and, inappropriate transformed clothing in case of overlaps among different clothing parts i.e., sleeves overlap on the torso (second row). 
	}
	\label{fig: tps_problem_demo1}
\end{figure}

\begin{figure}[!t]
    \centering
        \includegraphics[width=\linewidth]{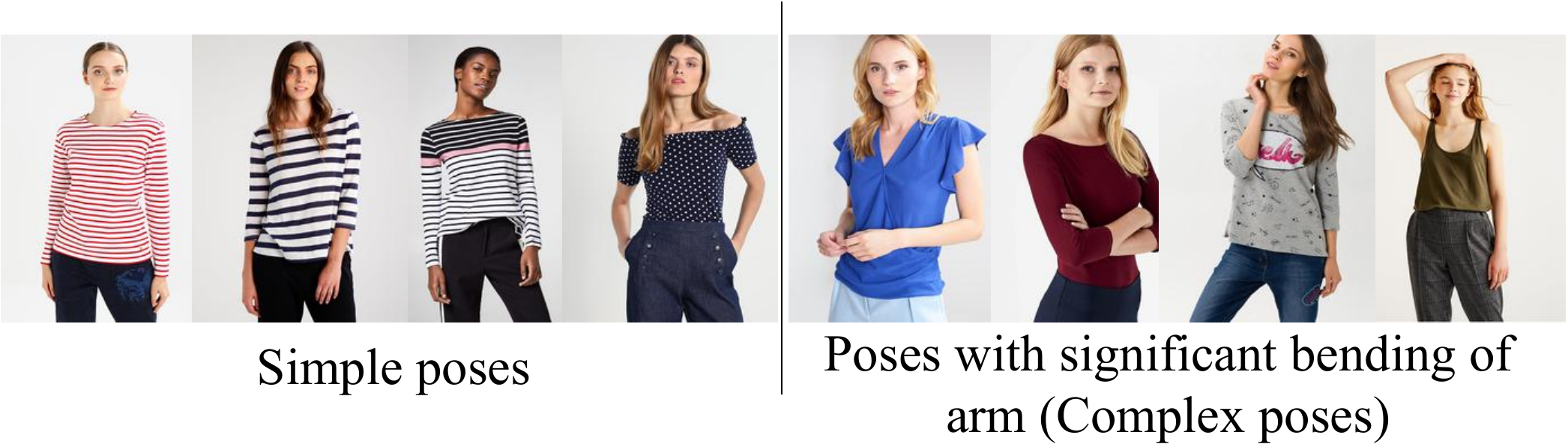}
 	\caption{Demonstration of simple and complex human poses.}
	\label{fig: demo}
\end{figure}

In terms of the solution approach, the existing VTON methods can be broadly classified into two genres: \emph{warping-based} and \emph{flow-based}. The warping-based methods~\cite{cpvton, vtnfp, lgvton, ivcnz_roy,wang2018non, cvpr_2020_2, multiposevton} predict a non-linear transformation, e.g., \emph{Thin Plate Spline (TPS) transform} to warp the source clothing in its target form. In the flow-based methods~\cite{chopra2021zflow, ge2021parser, han2019clothflow, he2022style} the target output is synthesized by predicting the appearance flow~\cite{zhou2016view} i.e., the vectors specifying the flow of pixels from the source to the target person. However, both types of approaches show poor performance when significant arm bending is observed as shown in Fig.~\ref{fig: tps_problem_demo1}. We name such postures as complex human postures as portrayed in Fig.~\ref{fig: demo}. In warping-based approaches, this mainly occurs due to the smoothness constraint of TPS which restricts bending in the predicted warp. To visually clarify, in Fig.~\ref{fig: tps_problem_demo1} we take examples of two benchmark methods one is flow-based (He et al.~\cite{he2022style}) and the other is warping-based (ACGPN~\cite{cvpr_2020_2}) and show that both these types of methods fail to warp the clothing in case significant bending is required in the target warp.

\IEEEpubidadjcol
As clothes are mostly designed based on human anatomy, different parts of clothing may move independently of each other. For instance, sleeves move independently of the torso part, just as our arms move independently of the torso. This independence may result in overlap between the different parts in case of folded or crossed-arm poses. Most of the previous warping-based methods employ TPS transformation which computes a single global transformation considering the clothing to be a single planar object. Hence, it disregards the independence of the movement of different clothing parts. The issue of overlaps is also observed in the results of flow-based methods, however, analyzing the cause requires understanding the network-learned features which is difficult. An example showing the issue of overlap occurring in both flow and warping-based methods is illustrated in Fig.~\ref{fig: tps_problem_demo1}.

In terms of the source clothing type, VTON methods can be categorized into two types: model-to-person (M2P) and cloth-to-person (C2P). C2P methods~\cite{viton, cpvton, multiposevton, cvpr_2020_2,he2022style, vtnfp} are trained on paired data, i.e., a person image with its corresponding in-shop clothing image ($C$) (shown in Fig.~\ref{fig: tps_problem_demo1}) as the source clothing. However, collecting such data is laborious~\cite{xie2021towards}. Whereas M2P methods~\cite{lgvton, ivcnz_roy, xie2021towards,han2019clothflow} take the image of a person say, model $M$, wearing the source clothing and the person image. These methods are generally trained on unpaired data using self-supervision. In addition to the challenges associated with learning with unpaired data, due to the pose difference between the model and person, an occluded clothing part in the source might become visible in the target, requiring the lost details to be predicted in the target. Hence, occlusion is a critical situation to deal with in M2P methods, but this does not occur in the C2P genre for obvious reasons.

Considering all these, we attempt to achieve a pose-robust M2P VTON solution that majorly contributes in two aspects. One, we propose a part-based warping approach that attempts to handle the cases of overlap among different clothing parts arising in crossed or folded arm postures; and two, considering the bending restriction imposed by the second order smoothness constraint of TPS that mainly affects the bending of sleeves in complex postures, we propose a geometric warping approach.

The proposed part-based warping tackles the overlap issue by employing clothing part-specific transformations. The idea of landmark~\cite{openpose} registration-based warping employing TPS transform proposed in LGVTON~\cite{lgvton} facilitates clothing transformation following human anatomical constraints; which aids in producing a more accurate target warp~\cite{lgvton}. We leverage this idea in warping the torso part of the clothing. However, warping the sleeves in case of significant arm bending is difficult using the TPS transform, due to its bending limitations. So, inspired from \textit{field transform}~\cite{beier} we propose an \emph{AnaTomy-Aware Geometric (ATAG) transform} that is consistent with human arm movements. Instead of landmark correspondences, here we consider the correspondences of the straight line segments between the consecutive landmarks of a human arm. This imposes an additional constraint that is well justified with human anatomy (see Fig.~\ref{fig: warping_blockdia}(b)) facilitating better warps.

The inherent challenge of occlusion is one of the major concerns in M2P methods. Our method tackles this using the proposed \emph{Mask Prediction Network (MPN)}. MPN predicts the target clothing mask that aids in distinguishing the areas of the target clothing occluded in the source. The proposed \emph{Image Synthesizer Network (ISN)} interpolates these occluded regions and also produces a seamless try-on image.

To summarize, we make the following contributions,
\begin{enumerate}
\item We propose a self-supervised model-to-person VTON solution that significantly improves performance over the previous methods, especially for complex human arm poses (see Fig.~\ref{fig: demo}). 
\item We propose a part-based solution approach that addresses the overlap issues by warping the independently movable parts separately.
\item We discuss the limitations of the TPS transform in the current problem context and propose a hand-crafted feature-based warping method (ATAG transform) that is consistent with the geometry of the human arm. 
\item Our proposed Mask Prediction Network (MPN) aids in identifying the occluded regions of the target clothing in the source and thereby guides the proposed Image Synthesizer Network (ISN) to predict lost clothing details in those regions.
\end{enumerate}
Our work is focused on upper-body clothing only; however, this can be extended to lower-body clothing too. In the rest of the paper, we present a brief literature survey in Sec.~\ref{related_works}, details of ATAG transformation in Sec.~\ref{approach_to_solution}, and the methodology in Sec.~\ref{methodology}. The experiments are discussed in Sec.~\ref{experiments}. Finally, we conclude in Sec.~\ref{sec: conclusion}. More results and code is available at: \url{https://dp-isi.github.io/ATAG/}
\section{Related Works}
\label{related_works}
Image-based VTON approaches~\cite{viton, cpvton, fitme, VTONnew, fashionon, coherence, m2e, vtnfp, multiposevton, fwgan, cvpr_2020_1, cvpr_2020_2, sievenet, han2019clothflow, 2018_CVPR_apptrans} have gained significant research attention over the past few years. The existing VTON methods can be categorized from different perspectives. Based on the solution approach VTON methods may be categorized into two genres - warping-based~\cite{viton, cpvton, vtnfp, lgvton, ivcnz_roy, cvpr_2020_2, multiposevton} and flow-based~\cite{chopra2021zflow, ge2021parser, han2019clothflow, he2022style}. Likewise, in terms of the clothing input we categorize into two - model-to-person (M2P)~\cite{lgvton, ivcnz_roy, xie2021towards, ge2021parser,han2019clothflow} and cloth-to-person (C2P)~\cite{viton, cpvton, multiposevton, cvpr_2020_2,he2022style, vtnfp, chopra2021zflow, ge2021parser}. Below we discuss the first type of categorization because of our emphasis on warping methodology.
Other than these some GAN-based image synthesis approaches for attribute manipulation related to VTON are proposed in~\cite{yildirim2019generating, men2020controllable, lewis2021tryongan, sievenet, garmentgan}.

\emph{Warping-based approaches -}
Among the warping-based approaches, the popular warping function is the Thin Plate Spline (TPS) transformation function. The warping-based approaches~\cite{cpvton, vtnfp, lgvton, ivcnz_roy,wang2018non, cvpr_2020_2, multiposevton,xie2021towards} employing deep neural networks learns the features from the clothing and the reference person images and correlates them. The features learned from that correlation map are used to predict the parameters of TPS. The network performing this is called the geometric matching network (GMN). CP-VTON~\cite{cpvton} first employed GMN and showed improvement over the shape context warp idea of VITON~\cite{viton}. VTNFP~\cite{vtnfp} additionally incorporates non-local mechanism~\cite{wang2018non} in the feature extraction part of the GMN. Unlike CP-VTON, MGVTON~\cite{multiposevton} learns the features from the mask of the clothing and the predicted target warp to 
capture only geometric features excluding texture and 
color information. However, the high flexibility of TPS, 
often causes GMN to produce undesirable warping results in the presence of complex patterns in clothes. ACGPN~\cite{cvpr_2020_2} proposed to employ a second-order difference constraint to control the grid deformations causing undesirable warps. VITON-HD~\cite{vitonhd} employs GMN and tackles the misalignment issues better with their proposed ALIAS normalization-based generator to effectively synthesize VTON results at high resolution. LGVTON~\cite{lgvton} proposed a landmark-guided VTON approach which used a landmark-based image registration approach to estimate the parameters of TPS. The work of Roy et al.~\cite{ivcnz_roy} employed GMN but learned the correspondences of features of the densepose representations~\cite{densepose} of the model and the person instead of directly learning from the images. The warping approach of C-VTON is very similar to that of Roy et al. The difference is that being a cloth-to-person approach here instead of the model's densepose the flat clothing source is given as input. Also, a powerful image generator is introduced. The key contribution of PASTA-GAN~\cite{xie2021towards} is employing the source and target person's pose-based key points to decouple a garment into normalized patches and then reconstruct them to the warped garment. While we also employ pose key points, the idea of our geometric warping approach (ATAG transform) is completely different.

\emph{Flow-based approaches -}
Apart from warping-based approaches, appearance flow-based approaches are also proposed. The very first work in this direction is proposed in~\cite{han2019clothflow}, which predicts the clothing flow from the source to target images. The main idea of appearance flow~\cite{liu2019liquid} is to predict the sampling grid for clothes warping. In PF-AFN~\cite{ge2021parser} Ge et al. propose a ``teacher-tutor-student" knowledge distillation approach to get rid of parsing information at inference time. SDAFN~\cite{sdafn} improves on it further by getting rid of the parsing information even during training and requiring only pose information. Consequently, a styleGAN~\cite{karras2019style} based appearance flow estimation method proposed by He et al.~\cite{he2022style} uses the style vector to capture the global context of the image to overcome the challenges arising from the consideration of only local appearance flow~\cite{chopra2021zflow, ge2021parser, han2019clothflow}. 
GP-VTON~\cite{gpvton} proposes a Local Flow Global-Parsing (LFGP) warping module that attempts to learn diverse local deformation fields for warping different garment parts. It also employs a gradient truncation strategy to make the warped garment align better with the preserved clothing details. While the previous approaches have shown commendable performance, but, in most cases struggle to warp the clothing in case of significant bending. For instance, in the case of folded or crossed-arm postures, the sleeves of the long-sleeved clothing require significant bending. The main contribution of this paper targets this limitation of others. Our method works on a variety of human poses, irrespective of human body shape, and outfit shapes, and also attempts to retain the exact texture and colors of the source clothing in the target output.
\section{Approach to Solution}
\label{approach_to_solution}
Existing transformation-based methods consider clothing as a single planar object that gets deformed. But as we know sleeves of the upper body clothing can move independently of the torso part causing overlaps among different parts. Ignoring this fact is likely to result in unrealistic transformation of clothes. So, in our proposed method we independently transform each part of the upper body clothing: torso, left sleeve, and right sleeve. Although the sleeves may be divided further into upper and lower parts for independent transformation, we are transforming the sleeve as a whole to get a realistic-looking warp near the elbow region. Since the movement of the sleeves solely depends on the posture of the arms, their transformation gets limited by the way we move our arms. To make the cloth transformation more realistic, we devise a transformation that takes into account the constraints of our arm movement.

A human arm consists of 3 bones: the humerus, radius, and ulna, apart from the bones in our hands. Since the two bones of the lower arm move together, we can consider them as one unit for our problem. Hence, we have two parts of our arm that are joined at the elbow, which constrains their movement. So, we model our transformation as the transformation of two straight lines joined at a point. Instead of a point-based transform, we propose the idea of a line-based transform that justifies the representation of the human arm as a set of bones instead of only the bone joints i.e., landmarks. Below we discuss our proposed line-based transform called AnaTomy-Aware Geometric (ATAG) Transform. 

\begin{figure*}[!t]
	\centering
	\includegraphics[width=\linewidth]{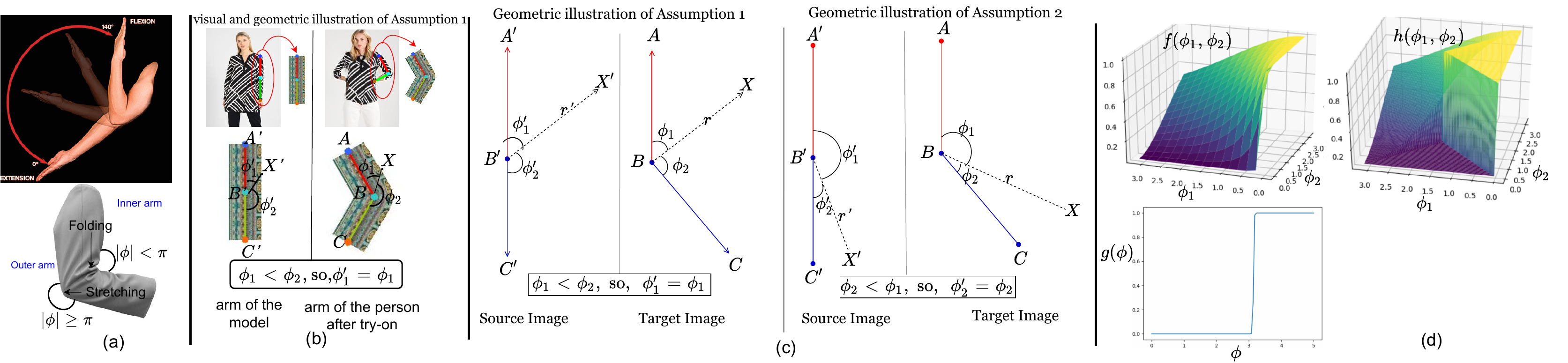}
	\captionof{figure}{(a)~(top) Elbow flexion and extension (picture courtesy~\cite{elbow_flexion}), (bottom)~Stretch and folds in clothing sleeve due to elbow flexion i.e., arm bending (picture courtesy~\cite{sleeve_folding}). (b) A graphical illustration of the arm-bending phenomenon in humans. A sample pair of model and person relevant to the illustrated scenario is given above for better understanding, (c)~Plot of functions $f(\phi_1, \phi_2)$, $g(\phi)$ and $h(\phi_1, \phi_2)$, (d)~Geometrical illustration of our warping method for sleeves warping for the two different scenarios. (Left)~Example of a case when assumption 1 holds. (Right)~Example of a case when assumption 2 holds. Here $\{A, B, C\}$ and $\{A', B', C'\}$ are the landmarks corresponding to the arm of the person and the model respectively. $X$ refers to the point belonging to the sleeve segment of the target warp and $X'$ is its corresponding source pixel.}
	\label{fig: all_2}
\end{figure*}

Before we discuss the ATAG transform, let us first denote $A'$, $B'$, $C'$ as the landmarks of an arm of the model and $A$, $B$, $C$ as the corresponding landmarks of the same arm in the target person. We compute the landmarks using the landmark localization approach proposed in~\cite{openpose}.
\subsection{AnaTomy-Aware Geometric (ATAG) Transform}
\label{atag_transform}
Let $AB$ and $BC$, connected at $B$, denote the upper and lower part of the arm in the target configuration and $A'B'$ and $B'C'$ denote its corresponding source configuration. We wish to compute a transformation such that each part of the arm influences only its nearby region. To compute this influence we utilize the polar coordinates of the locations with respect to the line segments. As shown in Fig.~\ref{fig: all_2}(b, c), let $X$ be a point in the target configuration and $X'$ be its corresponding location in the source configuration. Let, $X=(r_1, \phi_1)$ be the corresponding polar co-ordinate where $r=||BX||$ and $\phi_1=\angle{XBA}$. Similarly, let $X'=(r'_1, \phi'_1)$. Considering the advantages of backward transform over forward transform, here, we compute a backward transform, i.e. for each target $X$, we compute its location in the source. More precisely, from which source position the target need to be filled. For that we need to compute the $(r'_1, \phi'_1)$ from $(r_1, \phi_1)$. Below, we first discuss the method to compute the angular coordinate ($\phi_1'$) followed by discussing our idea of computing the radial coordinate ($r'$). Now before continuing further, let us introduce some notations. Let $\phi_2$ = $\angle{CBX}$, $\phi_1'$ = $\angle{X'B'A'}$. $\phi_2'$ = $\angle{C'B'X'}$, $\phi$ = $\angle{CBA}$ = $\phi_1 + \phi_2$.

\subsubsection{Computing the angular coordinate ($\phi_1'$)}
In general human arm undergoes bending which is called \textit{elbow flexion}. The bending amount is quantified using the \textit{flexion angle} ($\in [ 0^{\circ} \text{(no bending)}, 145^{\circ} \text{(max bending)}]$)~\footnote{https://en.wikipedia.org/wiki/Elbow}(Fig.~\ref{fig: all_2}(a)). Due to bending the sleeve undergoes folding in the inner part and stretching in the outer part (Fig.~\ref{fig: all_2}(a)). We observed, due to bending, the relative angular position of a source pixel $X'$ (target pixel $X$) with its closest line remains unchanged in the target warp (source clothing) due to the independent nature of the movement of the pixels around the upper and lower arm, except at the elbow region. For instance, as shown in Fig.~\ref{fig: all_2}(b), X is closer to $BA$ than $BC$ in terms of angular distance as $\phi_1 < \phi_2$ so $\phi_1' = \phi_1$. We include this observation in the form of two assumptions,
\begin{itemize}
	\item Assumption 1: When $X$ is closer to the line $BA$ i.e., when $\phi_1 < \phi_2$,
	the relative angular position of $X$ with respect to line $BA$  i.e., $\phi_1$, will be equal to the angular position of $X'$ with respect to line $B'A'$ which is $\phi_1'$, i.e., $\phi_1' = \phi_1$.
	\item Assumption 2: In a similar sense, when $X$ is closer to the line $BC$ i.e., when $\phi_2 < \phi_1$, then $\phi_2' = \phi_2$. Note that saying $\phi_2' = \phi_2$ is equivalent to saying $\phi_1' = \phi' - \phi_2$.
\end{itemize}
In the elbow region bending causes 2 kinds of warping of the sleeves, folding and stretching as shown in Fig.~\ref{fig: all_2}(a). 

In the outer arm area near the elbow region where generally $\phi_1$ and $\phi_2$ are very close, strictly adhering to one of two assumptions does not produce realistic results. Instead, then we compute $\phi_1'$ as the weighted combinations of the values of $\phi_1'$ computed by the two assumptions; where the weight is computed by a function $f(\cdot, \cdot)$. However, as $X$ moves towards the upper and lower end of the arm, either of the two assumptions can be used to compute $\phi_1'$. Based on this we compute $\phi_1'$ for the pixels in the outer arm region as,
\begin{equation}
	\phi_1' = \phi_1 (1 - f(\phi_1, \phi_2)) + (\phi' - \phi_2) f(\phi_1, \phi_2),
	\label{eq: theta1prime_1}
\end{equation}
where, 
\begin{equation}
    f(\phi_1, \phi_2) = \frac{\phi_1^2}{\phi_1^2 + \phi_2^2}
    \label{eq: func_f}
\end{equation}
The choice of this function is intuitive. Observe in Fig.~\ref{fig: all_2}(d) that the functional value of $f$ takes smooth transition from 0 to 1 from the region $\phi_1 < \phi_2$ (criterion of assumption 1) to $\phi_1 > \phi_2$ (criterion of assumption 2). This smooth transition ensures that near the elbow in the outer arm area i.e. when $\phi_1$ is close to $\phi_2$, a weighted combination of the effects of both the assumptions hold for stretching, unlike the inner arm.

For the sleeve area in the inner arm region, either of the two assumptions holds. A rounded version of the function $f$ may be used in such cases because rounding removes the effect of smooth transition which is not required here. So for the pixels in the inner arm region, we compute $\phi_1'$ as, 
\begin{equation}
	\phi_1' = \phi_1 (1 - Round(f(\phi_1, \phi_2))) + (\phi' - \phi_2) Round(f(\phi_1, \phi_2)).
	\label{eq: theta1prime_2}
\end{equation}

To combine these two cases in one expression we need a function to distinctly identify whether a pixel belongs to the inner arm or outer arm. Notice in the inner arm $\phi < \pi$ and in the outer arm $\phi \geq \pi$ (Fig.~\ref{fig: all_2}(a)). Based on this we define $g(\cdot, \cdot)$ as a logistic function that varies from 0 to 1 near $\pi$. The idea of taking a logistic instead of a step function here is to make a smooth transition between the regions $\phi < \pi$ and $\phi > \pi$ so that a realistic warping result is obtained. The functional form of $g$ is,
\begin{equation}
    g(\phi) = \frac{1}{1 + e^{a(\pi - \phi)}}.
    \label{eq: func_g}
\end{equation}
Observe, when $\phi < \pi$ (inner arm), $g(\cdot, \cdot) \to 0$, and, when $\phi > \pi$ (outer arm), $g(\cdot, \cdot) \to 1$. Now we know, in the outer arm $f(\phi_1, \phi_2)$ holds and in the inner arm $Round(f(\phi_1, \phi_2))$. Combining these two using $g(\cdot, \cdot)$ we get $h(\cdot, \cdot)$ as follows, 
\begin{equation}
	h(\phi_1, \phi_2) = g(\phi) f(\phi_1, \phi_2) +  (1 - g(\phi))Round(f(\phi_1, \phi_2)). 
	\label{eq: func_h}
\end{equation}

Hence, for all pixel $X$, we may compute $\phi_1'$  as,
\begin{equation}
	\phi_1' = \phi_1 (1 - h(\phi_1, \phi_2)) + (\phi' - \phi_2) h(\phi_1, \phi_2),
	\label{eq: theta1prime}
\end{equation}

\subsubsection{Computing the radial coordinate ($r'$)}
\begin{figure}
	\centering
	\includegraphics[width=0.45\linewidth]{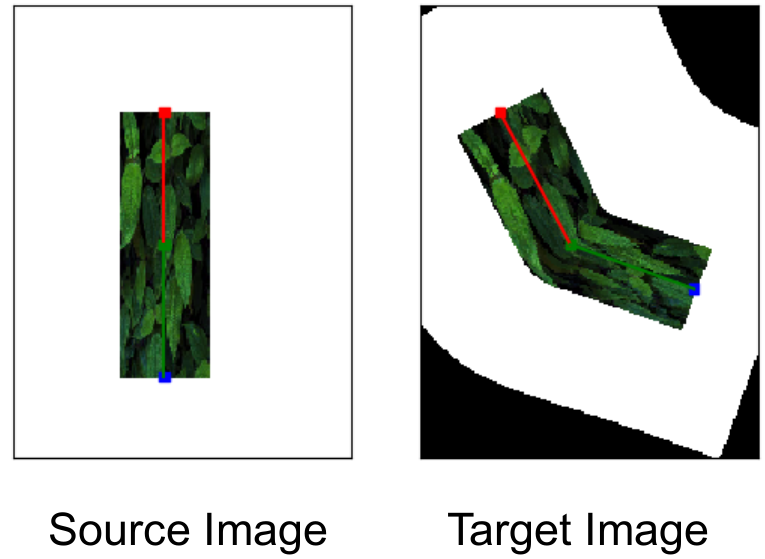}
	\captionof{figure}{Our result depicts the effect of scaling of lines in our method. As the line length increases from source to target the corresponding part of the rectangle looks zoomed-in in the result.}
	\label{fig: warp_demo2}
\end{figure}

Now that, we have computed $\phi_1'$, we need to compute $r'$. Before we compute $r'$ let us understand that a point closer to $B'C'$ (or $B'A'$) in the source should be proportionally similar in distance to $BC$ (respectively $BA$) in the target. To maintain this scaling effect we multiply $r$ with the ratio of the source and the corresponding target line i.e., $\|B'C'\|$ and $\|BC\|$  (resp. $\|B'A'\|$ and $\|BA\|$). For example, if $\|B'A'\| < \|BA\|$ then, $r'< r$, as $\frac{\|B'A'\|}{\|BA\|} < 1$. A demonstration of this case is given in Fig.~\ref{fig: warp_demo2}. In this example, the line $BA$ (red-colored) in the target is longer than $B'A'$ in the source, hence, the upper part of the sleeve (rectangle) in the transformed image is scaled accordingly. Based on this concept, we obtain the value of $r'$ using the following formula. 
\begin{equation}
	r' = r\left\{(1 - h(\phi_1, \phi_2)) \frac{\|B'A'\|}{\|BA\|}  +  h(\phi_1, \phi_2) \frac{\|B'C'\|}{\|BC\|}\right\}.
	\label{eq: rprime}
\end{equation}
The concept of employing $h(\cdot, \cdot)$ in Eq.~\ref{eq: rprime} is the same as that discussed previously.

Now, we have computed the radial and angular coordinates $r'$ and $\phi_1'$ of $X'$ respectively and already have the coordinates of $A'$ and $B'$. Therefore, computing the location of $X'$ in Cartesian is straightforward.

The steps of the proposed ATAG transform are formally presented in Algorithm.~\ref{algo: 1}. 

\begin{algorithm}
\caption{ATAG Transform (See Fig.~\ref{fig: all_2}(b) for the notations). Here without loss of generality, the ATAG transform for the left sleeve is presented.}
\label{algo: 1}
\begin{algorithmic}[1]
\renewcommand{\algorithmicrequire}{\textbf{Input: \{$A', B', C'$\},\{$A, B, C$\}, $X \in c_{lsleeve}'$}} 
\renewcommand{\algorithmicensure}{\textbf{Output:} Pixel $X' \in c_{lsleeve}$ corresponding to $X$}
\REQUIRE
\ENSURE
\STATE Convert $X$ to $(r, \phi_1)$: $r \leftarrow \|BX\|$ and $\phi_1 \leftarrow \angle{XBA}$.
\STATE $\phi_2 \leftarrow \angle{CBX}$.
\STATE $ h \leftarrow g(\phi) f(\phi_1, \phi_2) +  (1 - g(\phi))$\texttt{round}$(f(\phi_1, \phi_2))$, where $\phi = \phi_1 + \phi_2$. (Eq.~\ref{eq: func_h})
\STATE $\phi' \leftarrow \angle{C'B'A'}$
\STATE $\phi_1' \leftarrow \phi_1 (1 - h) + (\phi' - \phi_2) h.$ (Eq.~\ref{eq: theta1prime})
\STATE $r' = r\left\{(1 - h(\phi_1, \phi_2)) \frac{\|B'A'\|}{\|BA\|}  +  h(\phi_1, \phi_2) \frac{\|B'C'\|}{\|BC\|}\right\}.$ (Eq.~\ref{eq: rprime})
\STATE Convert $(r', \phi_1')$ to $X'$ (Cartesian coordinate).
\RETURN $X'$ 
\end{algorithmic}
\end{algorithm}
\section{Methodology}

\label{methodology}
\begin{figure}
	\centering
	\includegraphics[width=\linewidth]{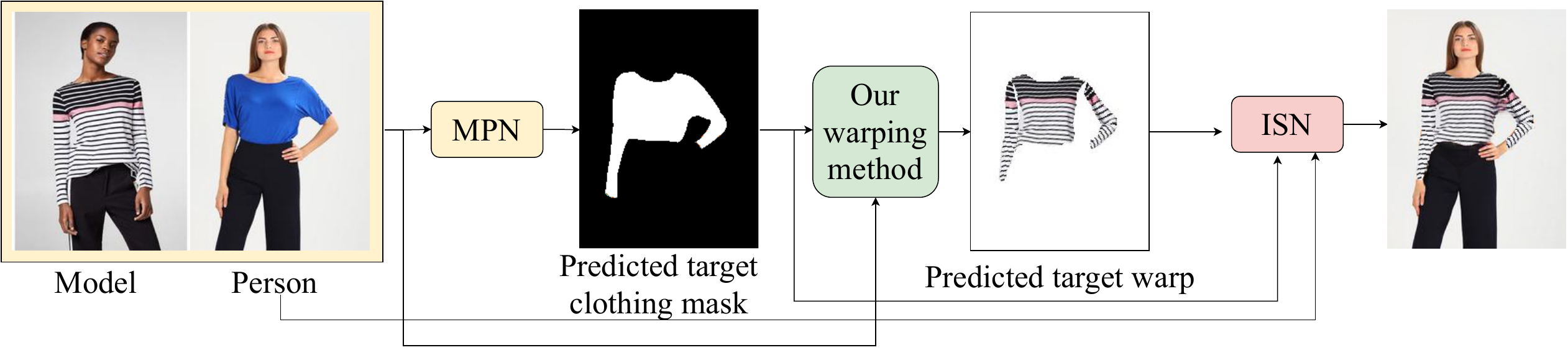}
	\captionof{figure}{Illustration of our overall virtual try-on approach.}
	\label{fig: overall_blockdia}
\end{figure}

Given a model's image $M$ and a person's image $P$
we wish to compute $P'$, which is, the image of the person wearing the reference clothing of $M$. We compute $P'$ in mainly three steps. First, We employ the proposed mask predictor network (MPN) that predicts the target clothing mask $\mathcal{S}$, denoting the trialed clothing region of the expected try-on output (Sec.~\ref{sec_mpn}). Second, we extract the model's outfit $c$ using a human parsing method~\cite{lipssl}\footnote{Human parsing is the task of segmenting a human image into 19 different fine-grained semantic parts e.g., head, torso, arms, legs, upper-body, lower-body clothes, etc.\label {humanParsing}} and compute the target warp $c'$ that fits the person $P$, using our proposed warping method (Sec.~\ref{sec_warp}). Third, we use our proposed image synthesizer network (ISN) that seamlessly combines $c'$ with $P$ to generate the final try-on output $P'$ (Sec.~\ref{sec_isn}). A simplistic block diagram demonstrating the different steps of our approach is shown in Fig.~\ref{fig: overall_blockdia}.

\subsection{Predicting the target clothing mask}
\label{sec_mpn}

\begin{figure*}[!htp]
	\centering
	\includegraphics[width=0.92\linewidth]{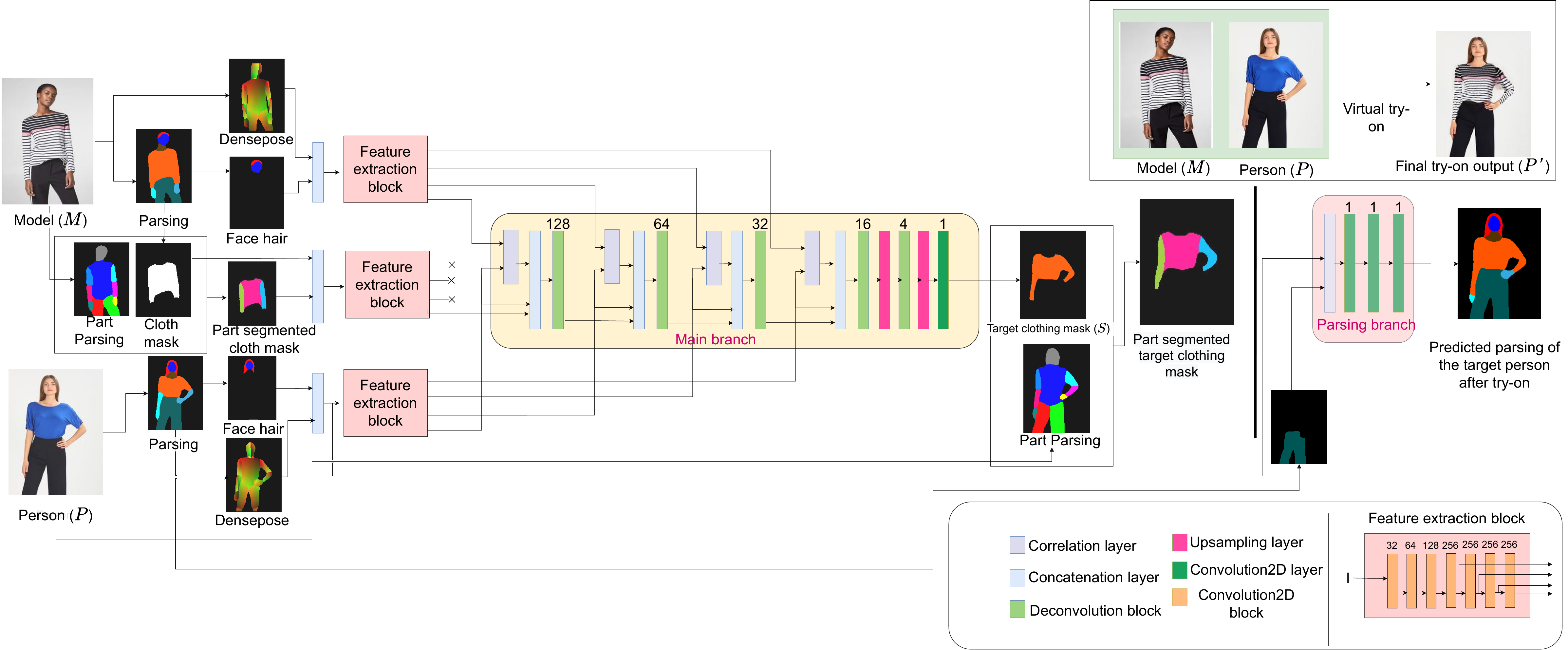} 
	\captionof{figure}{~Block diagram of the proposed mask predictor network (MPN). Best viewed in electronic version.}
	\label{fig: mpn_blockdia}
\end{figure*}

This step predicts the target clothing mask $\mathcal{S}$ that aids the synthesis module (ISN) to identify the occluded regions of the source clothing which needs to be interpolated to produce a seamless try-on output. To predict this mask, we train a convolutional neural network (CNN), named the mask prediction network (MPN), with the following inputs: (1)~the underclothing body shape of the model and the person, encoded using their densepose representations~\cite{densepose}~\footnote{A densepose representation maps all human pixels of an RGB image, to the 3D surface of the human body, thus providing a precise estimate of the human body shape under the clothing.}. In addition, we also provide the face and head segments of both the model and the person extracted using~\cite{lipssl}. (2)~the mask of the source clothing $c$, and (3)~semantically segmented human parts~\cite{cdcl} of the model. A block diagram of MPN along with the inputs and output are demonstrated in Fig.~\ref{fig: mpn_blockdia}. Note that providing the face and head segments along with the densepose as input gives the whole detail of a human body. Whereas, a more refined level of detail is provided by the segmented human parts.
 
The architecture of MPN is demonstrated in Fig.~\ref{fig: mpn_blockdia}. We incorporate correlation layers in MPN that compute the linear relationship between the body shape and pose features of the model and that of the person (Observe Fig.~\ref{fig: mpn_blockdia}). At the end of this module, the human parsing branch predicts the human parsing or semantic layout of $P'$ given the non-target clothing details of $P$ e.g., the lower body clothing, the face and hair segments, and the predicted mask from the previous branch. The role of this branch is to enforce a constraint on the main branch to predict a mask that is consistent with the final human parsing.
\subsection{Extracting the model's clothing and Predicting the target warp}
\label{sec_warp}

\begin{figure*}[!htp]
	\centering
	\includegraphics[width=0.95\linewidth]{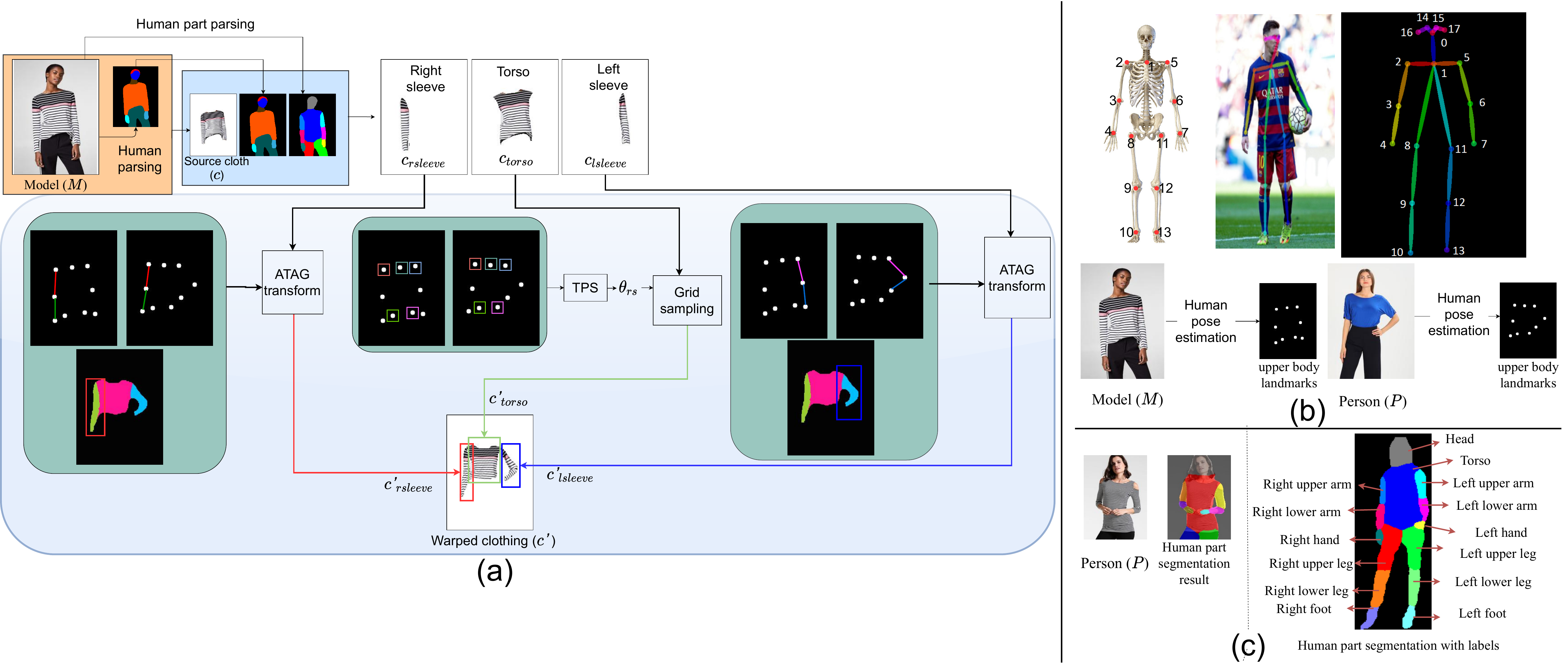} 
	\captionof{figure}{(a)~Illustration of the method for predicting the target warp, (b)~Demonstration of human landmarks, (c)~Demonstration of human part parsing~\cite{cdcl} results. Best viewed in electronic version.}
	\label{fig: warping_blockdia}
\end{figure*}

To extract the source clothing $c$ from the model image we utilize the human parsing method of~\cite{lipssl} which provides fine-grain segmentation of 19 different semantic parts\footref{humanParsing}. To get finer segments of $c$ such as right sleeve ($c_{rsleeve}$), left sleeve ($c_{lsleeve}$) and torso ($c_{torso}$), we opt for the human part-parsing approach~\cite{cdcl} demonstrated in Fig.~\ref{fig: warping_blockdia}(c). Such finer segmentation is utilized in our part-based transformation approach.

In our part-based warping approach, we separately compute the target warps $c_{torso}'$, $c_{lsleeve}'$ and $c_{rsleeve}'$ corresponding to each of the three source clothing parts, $c_{torso}$, $c_{lsleeve}$ and $c_{rsleeve}$ respectively. For computing $c_{torso}'$ from $c_{torso}$ we adopt the landmark-based warping method proposed in LGVTON~\cite{lgvton} which uses human landmark correspondences to compute the TPS transform~\cite{lgvton}. To compute $c_{lsleeve}'$ from $c_{lsleeve}$ and $c_{rsleeve}'$ from $c_{rsleeve}$ we use our proposed ATAG transform (Sec.~\ref{atag_transform}). The overall warping approach is illustrated in Fig.~\ref{fig: warping_blockdia}(a). Illustrations of human landmarks are given in Fig.~\ref{fig: warping_blockdia}(b).

Precisely, we use the ATAG transform to compute the source pixel corresponding to each pixel in the segments, $c_{lsleeve}'$ and $c_{rsleeve}'$, and the TPS transform to compute that for $c_{torso}'$. The target warp ($c'$)'s mask $c'_m$  = $\mathcal{S} - s$, where $\mathcal{S}$ is the target clothing mask predicted by MPN and $s$ contains the pixels $\in \mathcal{S}$ not having the corresponding source pixel in $c$. More specifically, $s$ denotes the pixels whose corresponding source pixels belong to the occluded regions of the source clothing which will be predicted by the proposed ISN (discussed later). We will discuss this in the next paragraph. Now to compute the mask of the finer segments of $c'$ i.e., $c_{lsleeve}'$, $c_{rsleeve}'$, $c_{torso}'$, we use the human part parsing output of $P$ providing the part segments left arm, right arm, torso, and the target warp's mask $c'_m$ (illustrated in Fig.~\ref{fig: warping_blockdia}(c)).

\subsection{Try-on image synthesis}
\label{sec_isn}

The objective of this stage is to combine the predicted target warp $c'$ with the person image $P$ to generate a plausible and photo-realistic try-on result. 

One of the main goals of ISN is to deal with occlusion. Our warping method computes the target warp $c'$ from only the visible clothing areas of the model's clothing. However, due to the difference in pose between the model and the person some occluded areas of the model's clothing might become visible in the target. Mainly, there are 3 causes of occlusion - (1)~arms in front of the clothing, (2)~folds in sleeves near the elbow (we call such occluded regions ambiguous regions), (3)~long hairs on clothing areas. (1), (3) can be easily identified. Though, finding (2) i.e., ambiguous regions is a bit tricky using the rules of ATAG transform it can be identified. We explain this further in the supplementary material. 

\begin{figure}[!h]
	\centering
	\includegraphics[width=0.95\linewidth]{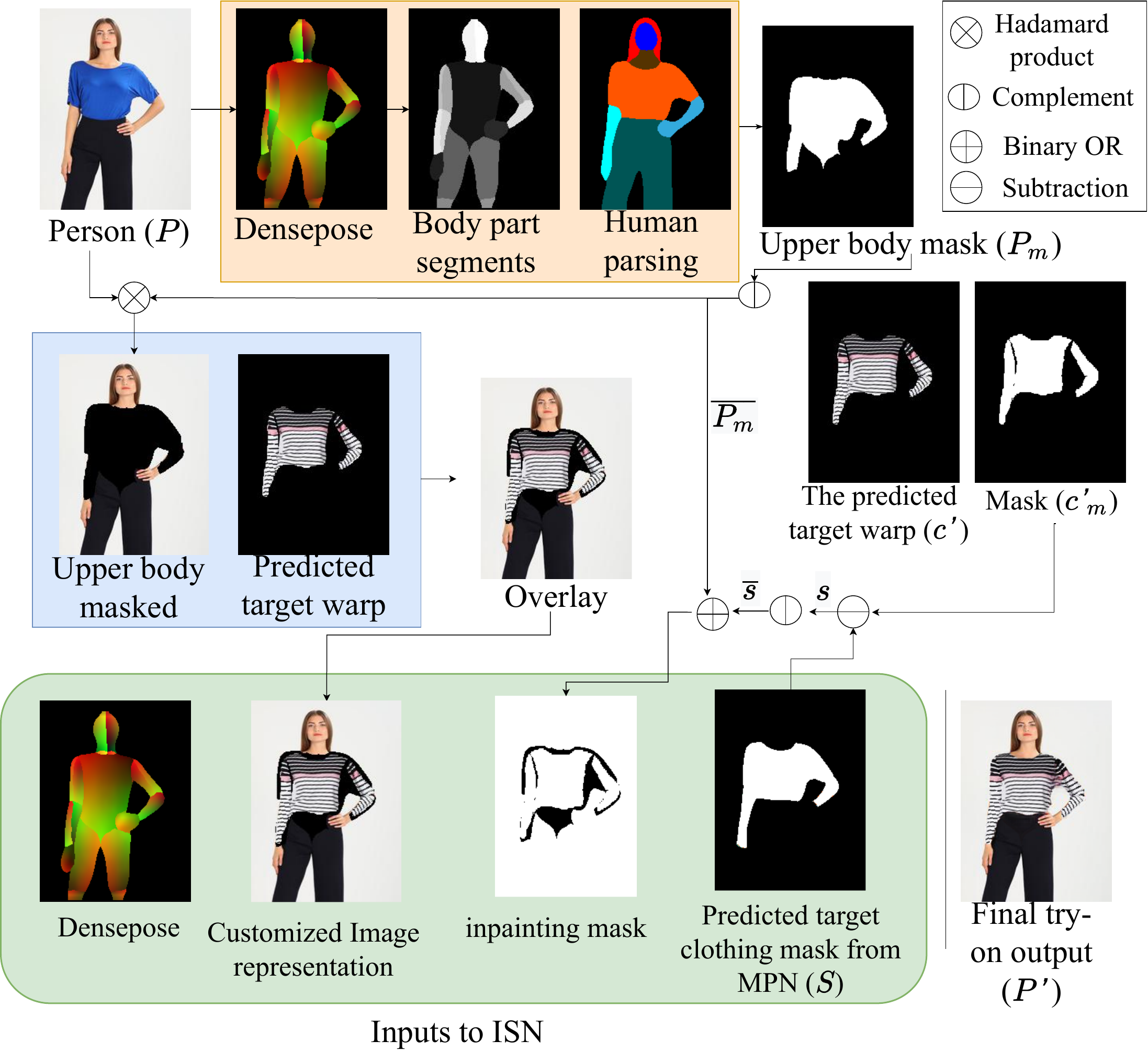}
	\captionof{figure}{Illustration of the inputs during test-phase to the ISN.}
	\label{fig: isn_inputs}
\end{figure}

Considering the upper body mask of the target person $P$ (including the clothing and skin areas) as $P_m$ (shown in Fig.~\ref{fig: isn_inputs}), the region to be predicted by ISN (inpaint mask) is computed as $\bar{P_m} \oplus \bar{s}$, where $\oplus$ denotes binary OR and $s$  = $\mathcal{S} - c'_m$. We call this the in-paint mask (0 - denotes the pixels to be predicted). A visual illustration of it is given in Fig.~\ref{fig: isn_inputs}. The objective of including $\bar{P_m}$ is to remove the previous clothing details from $P$. Now, the goal of ISN is to inpaint this region and produce a seamless combination of $c'$ and $P$. 

To achieve the above objectives we trained an encoder-decoder-based convolutional neural network (CNN). We call it the image synthesizer network (ISN). The inputs to this network are: (i)~the predicted target clothing mask from the MPN, (ii)~the body shape of the target person in densepose representation, (iii)~a combined representation of the predicted target warp and the person image, instead of providing them separately.
To compute this, the upper body details i.e., previous clothing (e.g., t-shirt, etc.) and exposed skin areas are masked in the person image, and the predicted warp is overlaid on it. (iv)~ The inpaint mask discussed previously.

We trained this network using self-supervision employing inpainting-related loss functions as suggested in~\cite{liu2018image}.

\section{Experimental Evaluation}
\label{experiments}
We compare our result with that of some benchmark try-on methods - 
CP-VTON~\cite{cpvton}, ACGPN~\cite{cvpr_2020_2}, MGVTON~\cite{multiposevton}, and He et al.~\cite{he2022style}, LGVTON~\cite{lgvton}~\footnote{Note that the results of LGVTON are reported based on human landmark annotations only because fashion landmarks are not available in the MPV dataset.}, Roy et al.~\cite{ivcnz_roy}, PASTA-GAN~\cite{xie2021towards}, C-VTON~\cite{cvton} and GP-VTON~\cite{gpvton}, VITON-HD~\cite{vitonhd}, SDAFN~\cite{sdafn} using two popularly used metrics: Fréchet Inception Distance (FID)~\cite{fid} and Structural SIMilarity index (SSIM)~\cite{ssim}. Note that, we compare with both M2P (~\cite{ivcnz_roy, lgvton, xie2021towards, ge2021parser}) and C2P (~\cite{cpvton, multiposevton, cvpr_2020_2,he2022style, cvton, gpvton, vitonhd, sdafn}) methods.

Now comparing the M2P methods directly with C2P methods is a bit unfair due to the additional advantages the C2P methods get in terms of the source clothing, such as - no occlusion, and no pose variability of the garment image. Also, C2P methods can't transfer the garment from model to person directly and hence can not be executed on M2P's input settings. Therefore, to introduce fairness in comparison we execute these methods on clothing segmented from the model's image in place of the garment image.\\
\textbf{Dataset.}
We conduct experiments on two benchmark datasets MPV~\cite{multiposevton} and VITON-HD~\cite{vitonhd} which mainly differ from each other in terms of resolution and viewpoint variation. While VITONHD contains images with higher resolution, images in MPV have higher viewpoint variations.

MPV contains 35,687 images (256 $\times$ 192) of fashion models captured from various angles. Their upper-body clothing images are given separately which is 13,524 in total. The test set - we made this set by randomly collecting 3000 image pairs. Note that this dataset does not contain any image of two different models wearing the same garment. Therefore, our test set does not have any ground truth images. We have two trainable networks MPN and ISN. The training set of MPN contains 56,968 pairs of same-person multi-view images collected from MPV where each pair contains different views of the same model's image. The training set of ISN - it contains 20,034 same-person image pairs i.e., the model and person images in each of the data samples are the same images. We chose MPV over some other benchmark datasets~\cite{deepfashion, viton} because DeepFashion~\cite{deepfashion} does not contain separate garment images, and MPV and VITON are collected from the same source website so they have similar data and even overlap in the contents. Also compared to VITON, MPV shows a higher degree of appearance and viewpoint variability.

VITON-HD contains 13,679 image pairs of front-view upper-body women and upper garments, which are further split into 11,647/2,032 training/testing pairs. We conduct experiments on this dataset at a resolution of ($512 \times 384$).

\subsection{Quantitative analysis}
\label{quantitative}
We quantitatively evaluate our results on two metrics - \emph{Fréchet Inception Distance (FID)}~\cite{fid}, the \emph{Structural SIMilarity metric (SSIM)}~\cite{ssim}. FID is used as a metric in the current problem context by previous many VTON methods~\cite{xie2021towards, cvpr_2020_1, ge2021parser, sievenet, garmentgan, lewis2021tryongan}. It measures the similarity between two sets of images based on the features extracted from a layer of inception v3 model~\cite{inceptionv3} pre-trained on ImageNet~\cite{imagenet}. 
A lower value of FID indicates better results. SSIM is a widely used image similarity metric, ranging in [0, 1] (1 indicates the same images).~\footnote{The detail on Fréchet Inception Distance (FID) is given in the supplementary materials.}

We present two quantitative studies on MPV in Table.~\ref{results_quantitative} and Table.~\ref{results_quantitative_pasta_gan}. Table.~\ref{results_quantitative} presents the results on both the test and the train set of MPV. As mentioned before, to introduce fairness in comparison we execute C2P methods on clothing segmented from the model's image in place of the garment image to report the scores presented in the M2P column (2$^{nd}$ column). The scores in C2P columns are on their usual input settings. Comparing the scores we see our method shows comparable performance with the benchmark methods achieving the best scores on the training set and second best on the test set.

Comparing with PASTA-GAN in Table.~\ref{results_quantitative} will not be fair as we could not reproduce their result. Hence, we compare with it in Table.~\ref{results_quantitative_pasta_gan} where we report our score on the test set reported in the PASTA-GAN paper~\cite{xie2021towards}. Observe that we perform better compared to others on this set. Note that, similar to ours both LGVTON and PASTA-GAN use model and person's pose key points in computing the target warp. But the methodologies are very different. 

Comparison on VITON-HD~\cite{vitonhd} is presented in Table.~\ref{results_quantitative_vitonhd}. Here, we compare one warping-based~\cite{vitonhd} and one flow-based parser-free method~\cite{sdafn} which have shown to perform well on high-resolution images. It is observed that our method performs better compared to both methods in M2P's input settings. Additional comparative studies on the VITON-HD dataset are given in the supplementary materials.

\begin{table}
\centering
	\caption{Quantitative evaluation on different datasets. In column 2 the inputs of C2P methods is ($c$, $P$) ($c$ is the clothing segment of $M$) and that of the others is ($M$, $P$). In column 5, C2P methods are run in their default inputs setting i.e., is ($C$, $P$). 
	In the training set as the source and the target images are the same, therefore no variation in pose, so, no question of occlusion, hence result will be the same for both clothing sources. The best and the second best results are highlighted with bold notation and blue color respectively.
	}
\begin{tabular}{c|c|cc|c}
\toprule
& Test Set  & \multicolumn{2}{c|}{Training Set} & Test Set\\

Method & (M2P) & & & (C2P)\\

& FID$\downarrow$ & FID$\downarrow$ & SSIM$\uparrow$ & FID$\downarrow$\\
 \midrule
MGVTON (ICCV, 19)~\cite{multiposevton} & 47.23 & 35.70 & 0.76 & 38.19\\
CP-VTON (ECCV, 18)~\cite{cpvton} & 50.67 & 21.03 & 0.74 & 40.02\\
LGVTON (MTA, 22)~\cite{lgvton} & 28.86 & 12.06 & 0.89 & - \\

Roy et al. (IVCNZ, 2020)~\cite{ivcnz_roy} & 28.41 & 14.34 & 0.80 & - \\

ACGPN (CVPR, 20)~\cite{cvpr_2020_2} & 27.80 & 14.35 & 0.88 & 17.87\\
He et al. (CVPR, 22)~\cite{he2022style} & 18.31 & 11.45 & 0.89 & \textcolor{blue}{9.38}\\
C-VTON (WACV, 22)~\cite{cvton} & \textbf{15.61} & \textcolor{blue}{10.28} & \textcolor{blue}{0.91} & \textbf{9.33}\\

GP-VTON (CVPR, 23)~\cite{gpvton} & 18.53 & 12.72 & 0.88 & 16.41\\
Ours & \textcolor{blue}{16.38} & \textbf{9.45} & \textbf{0.93} & - \\

\bottomrule
\end{tabular}
\label{results_quantitative}
\end{table}

\begin{table}
\centering
	\caption{The FID scores on the MPV test set used in PASTA-GAN. The scores of all the methods except ours are taken from the paper PASTA-GAN.
	}
\begin{tabular}{c|c}
\toprule

Method & M2P Test Set FID$\downarrow$ \\
 \midrule
CP-VTON (ECCV, 18)~\cite{cpvton} & 37.72\\
ACGPN (CVPR, 20)~\cite{cvpr_2020_2}& 23.20\\
PFAFN (CVPR, 21)~\cite{ge2021parser}& 17.40\\
PASTA-GAN (NeurlIPS, 21)~\cite{xie2021towards} & 16.48\\
Ours & \textbf{15.74} \\
\bottomrule
\end{tabular}
\label{results_quantitative_pasta_gan}
\end{table}

\begin{figure*}[!h]
	\centering
 \includegraphics[width=0.9\linewidth]{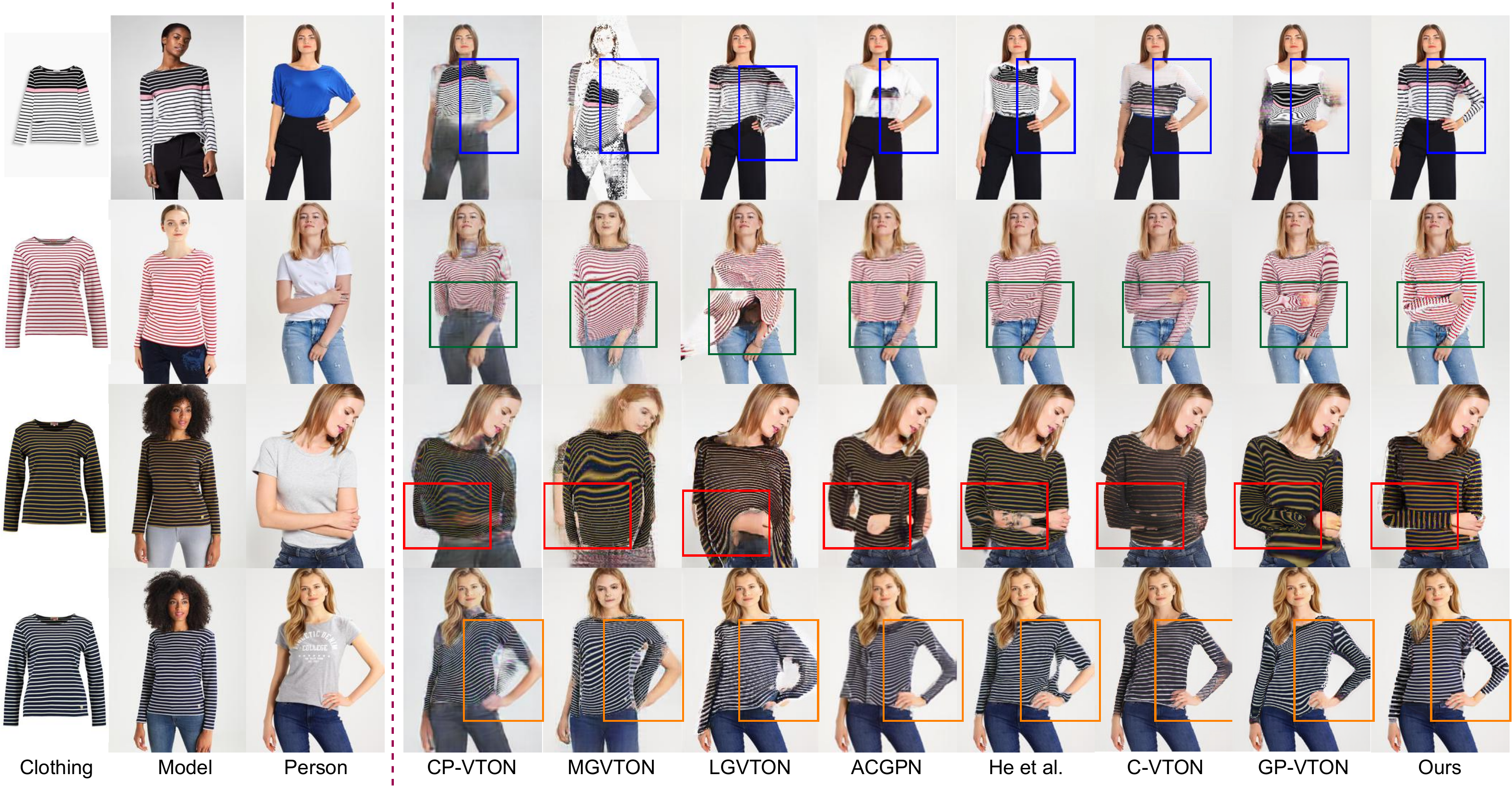}
	\captionof{figure}{Comparative study on MPV dataset in the cases of significant arm bending.}
	\label{fig: results_compare1}
\end{figure*}
\begin{figure}[!h]
	\centering
	\includegraphics[width=0.9\linewidth]{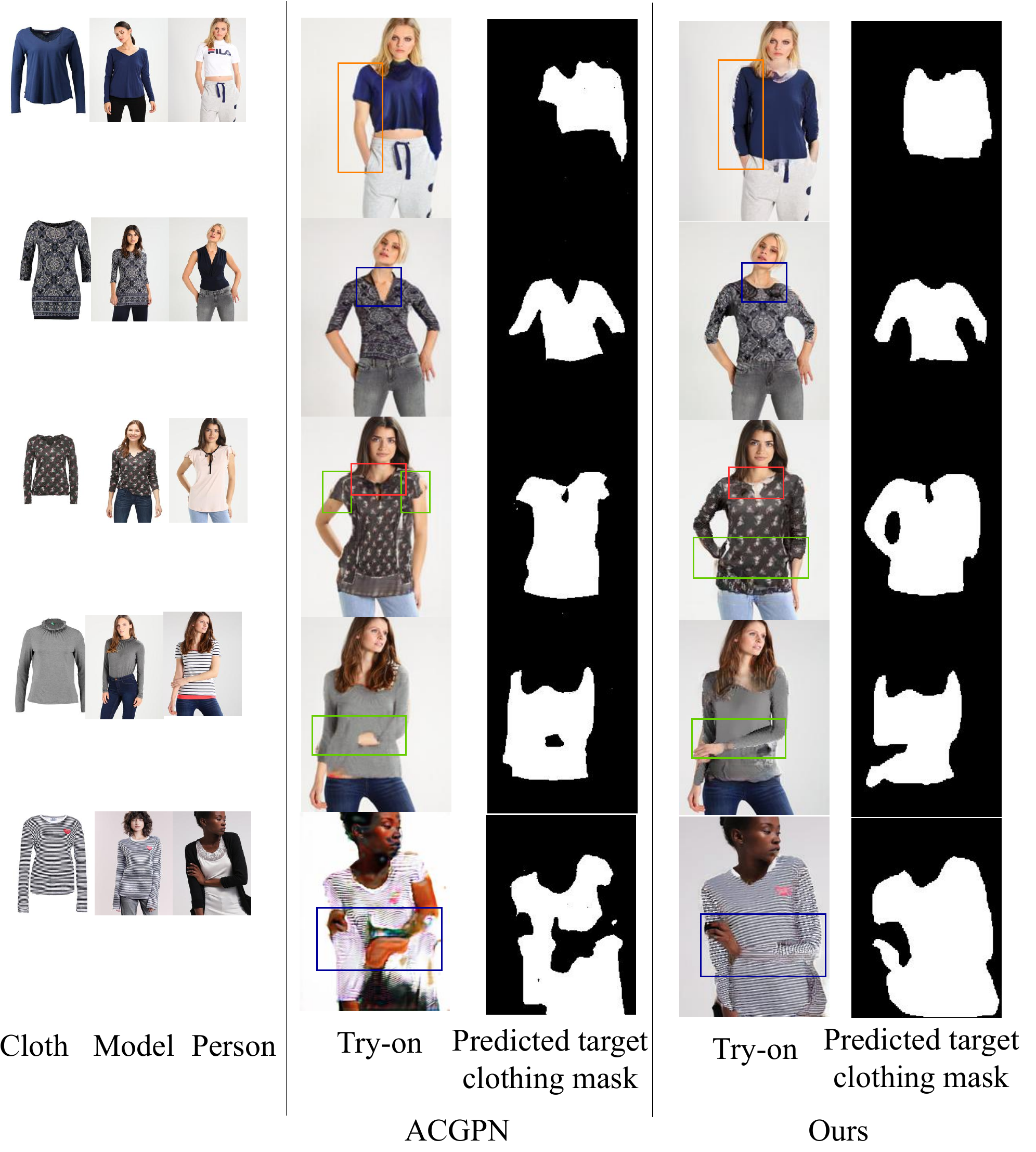}
	\captionof{figure}{Qualitative comparison of the proposed MPN with the semantic generation module (SGM) of ACGPN on the MPV dataset. We have highlighted the areas to be noticed in each of the results. Observe in most cases SGM does not preserve the features of the source clothing in the try-on result, instead keeps those features the same as the old cloth of the person (i.e., before try-on).}
	\label{fig: mpn_output1}
\end{figure}

\begin{table}[!h]
\centering
	\caption{The FID scores on the VITON-HD test set.}
\begin{tabular}{c|cc}
\toprule
\multirow{2}{*}{Method} & \multicolumn{2}{c}{Test Set FID$\downarrow$}\\
& M2P & C2P \\
 \midrule
VITONHD (CVPR, 21)~\cite{vitonhd} & 29.83 & 14.05\\
SDAFN (ECCV, 22)~\cite{sdafn} & 32.64 & \textbf{10.55}\\
Ours & \textbf{25.69} & $-$\\
\bottomrule
\end{tabular}
\label{results_quantitative_vitonhd}
\end{table}
\begin{figure}[h]
	\centering
	\includegraphics[width=\linewidth]{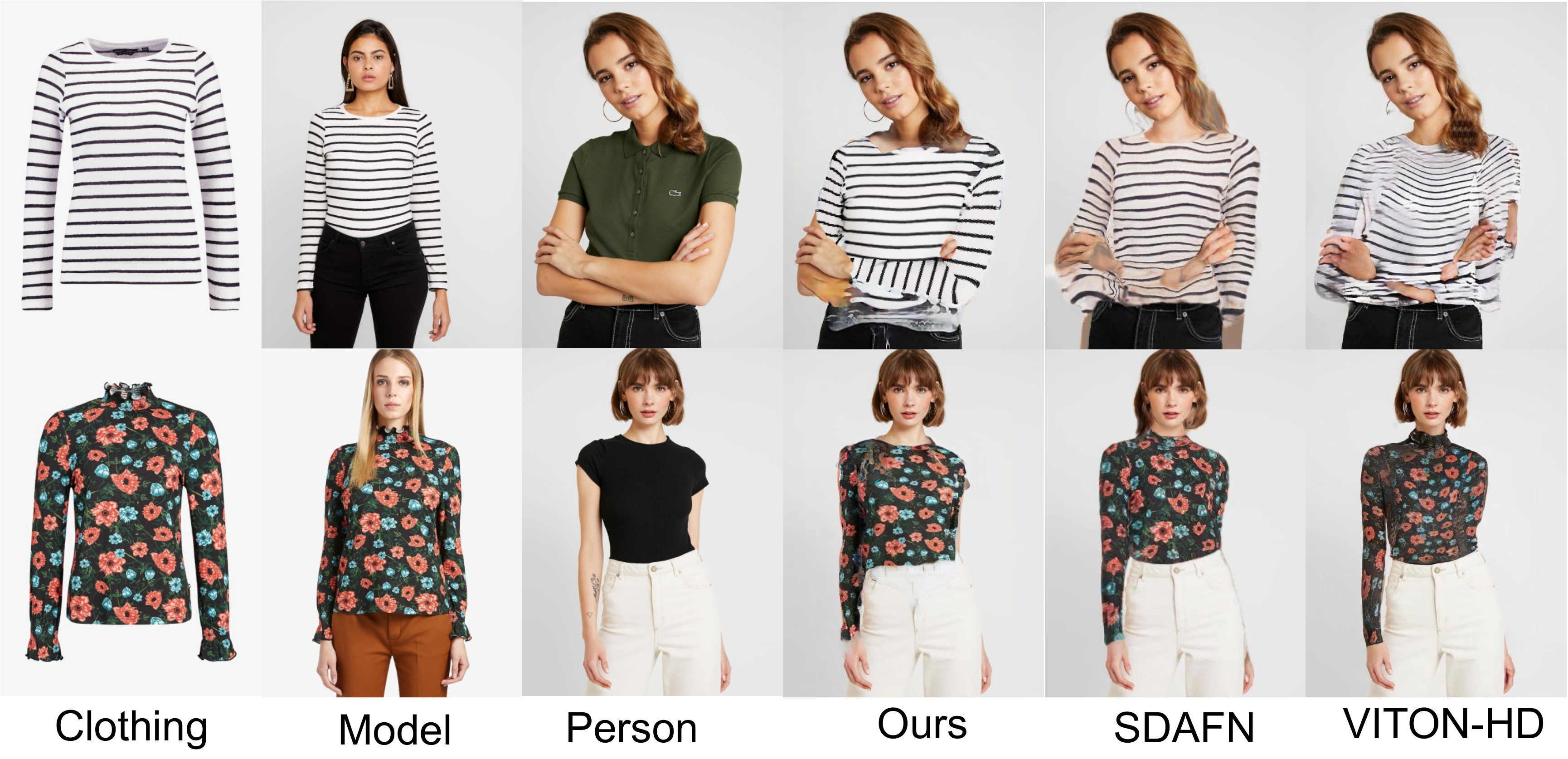} 
	\captionof{figure}{Results on VITONHD dataset. Please zoom in for a better view.}
	\label{fig: vitonhd_res1}
\end{figure}

\subsection{Qualitative Analysis}
Visual comparison with some benchmark methods on the MPV dataset is presented in Fig.~\ref{fig: results_compare1}. The results visually show the limitations of the C2P methods, especially in complex postures. Some additional comparative studies and results of our method on the different clothing and pose variety are given in the supplementary materials.
~\footnote{\label{footnote: qualitative}To compare all the methods to the best of their ability, all the results are generated in the default setting of the respective methods i.e., the garment image as input for the C2P methods and the model image to M2P methods.}

It can be observed that while handling critical hand bending all the methods except ours fail to perform satisfactorily. In the case of CP-VTON and MGVTON, the poor warpings are probably occurring due to the inappropriate learning of human body geometry by the geometric matching network. ACGPN shows some improvement due to its second-order difference constraint. He et al. perform well but still can not tackle near the areas of critical arm bendings. Notice the blurry areas in the try-on clothing regions of CP-VTON and ACGPN. These are occurring when these methods have tried to compensate for the limitations of their warping method, by filling some inappropriately warped regions in their try-on stage. However, the blurring is caused because the try-on synthesis module is not strong enough to fill the texture or color accurately. Although C-VTON and GP-VTON perform better visible degradation is observed with increasing bending of the arm (Observe the results in the second and third rows). Similar to ours, GP-VTON also employs the idea of warping the sleeves and torso separately. However, the superiority of the proposed ATAG transform over GP-VTON's flow-based warping approach is evident from the examples in Fig.~\ref{fig: results_compare1}.

Similar to LGVTON employing the idea of landmarks correspondence in warping produces correct deformation in the torso region. However, LGVTON's idea of considering whole clothing as a single planar object causes inaccurate warping of sleeves; which is tackled by the part-based approach that enables us to handle the cases of overlap between sleeve and torso. Unlike the others, we can handle the bending of the sleeves better due to our ATAG transform-based warping approach. 

We also present a visual comparison between the results of our MPN and the corresponding semantic generation module (SGM) of ACGPN in Fig.~\ref{fig: mpn_output1} because the objectives of these modules are the same (i.e. predicting target clothing mask). The results show some significant features (e.g., neck pattern, sleeve length, highlighted in Fig.~\ref{fig: mpn_output1}) of the target clothing mask are predicted better by our MPN compared to SGM.

Visual comparison of the VITON-HD dataset is presented in Fig.~\ref{fig: vitonhd_res1}~\footref{footnote: qualitative}. It can be observed that compared to both SDAFN and VITON-HD with increasing arm bending our method can warp the clothing better. This shows that the ATAG transform can better warp the clothing compared to both flow-based~\cite{sdafn} and TPS-based~\cite {vitonhd} warping approaches even in the case of high-resolution inputs. This is because the ATAG transform is learning-independent and does not get affected by the input resolution. However, it can also be observed in our result, that the fit of the try-on clothing and inpainting of occluded clothing details are not satisfactory (see the lower part of the try-on clothing in the output given in the first row). We analyze this further in Sec.~\ref{limitations}.

\begin{figure*}[!h]
	\centering
	\includegraphics[width=\linewidth]{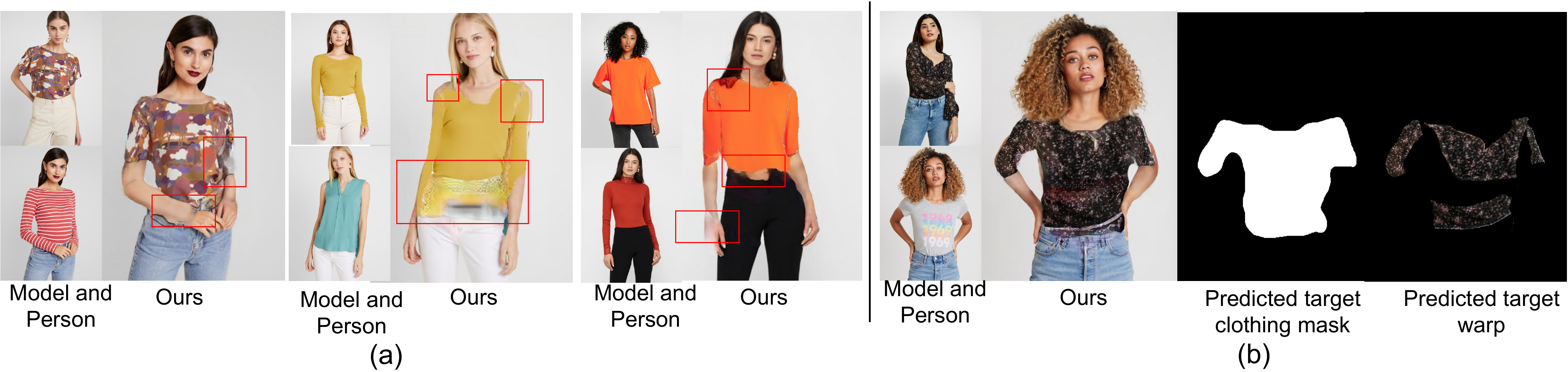}
	\captionof{figure}{(a)~Improper inpainting by the proposed ISN, (b)~Limitation of performance of MPN across datasets.}
	\label{fig: limitations}
\end{figure*}
\subsection{Ablation Study}
\label{ablation_study}

\begin{figure}[!t]
	\centering
	\includegraphics[width=\linewidth]{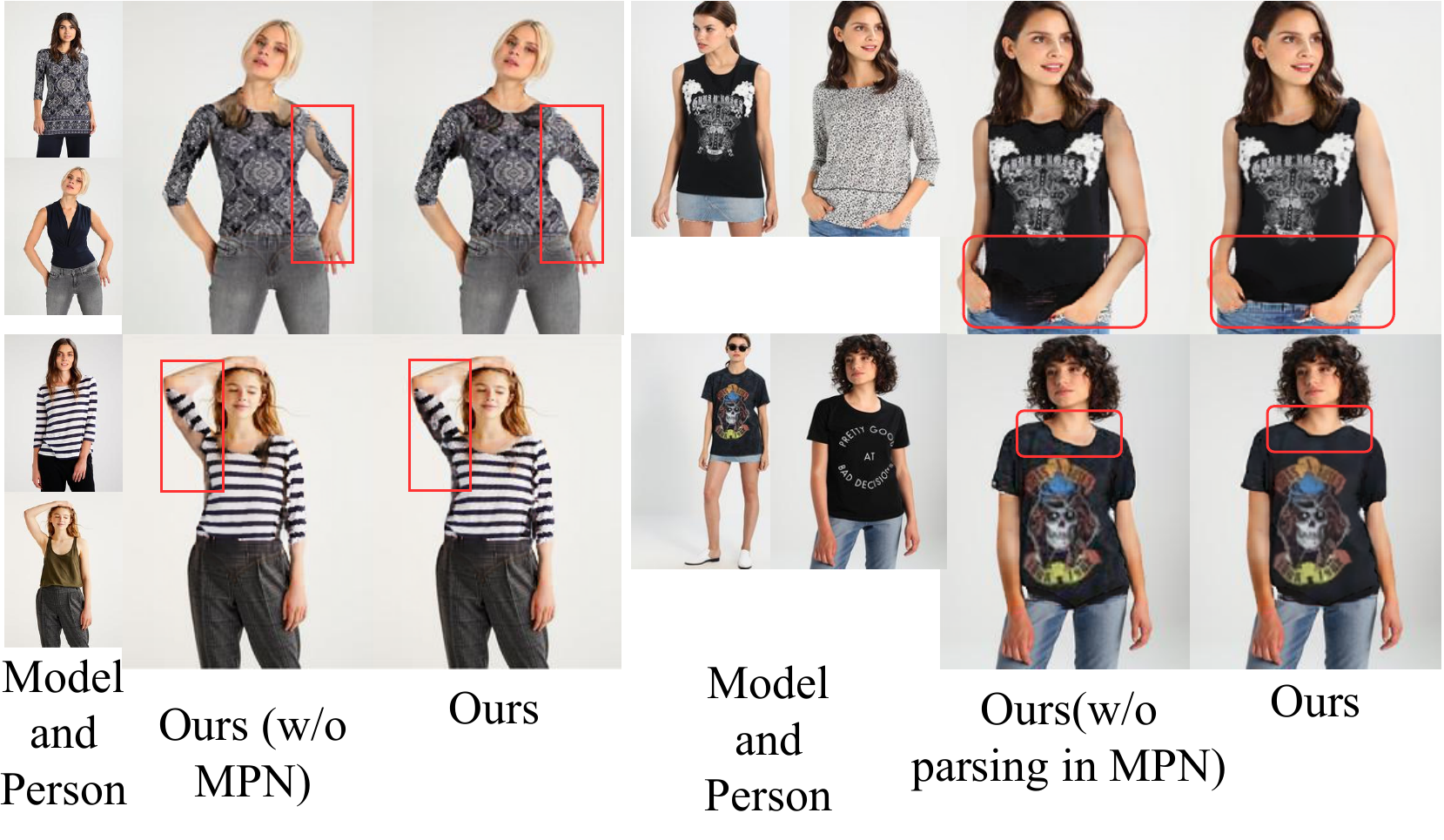}
	\captionof{figure}{Illustration of the significance of MPN with and without the parsing branch in the synthesis stage.}
	\label{fig: ablation2}
\end{figure}

We analyze the significance of employing MPN and its human parsing branch in our approach. In the former case, we compute the try-on results without using the mask predicted by MPN i.e., $\mathcal{S}$. Here, we compute the target warp's mask $c'_m$  = $P_m - s$, where $P_m$ is the upper body mask of $P$. In this case, we use $c'_m$ in place of $\mathcal{S}$ in subsequent stages. The difference it makes with our proposed approach is that the occluded parts of source clothing are not predicted in the output. The degradation in performance due to this is clearly noticeable in the highlighted region of Fig.~\ref{fig: ablation2} (Ours (w/o MPN)) and also verifiable quantitatively in Table.~\ref{tab: ablation2}.

For the latter case, we trained an instance of MPN without the parsing branch and computed the final results using that. Compared to the w/o parsing branch in the MPN instance, our method secures a better FID score (Table.~\ref{tab: ablation2}).  Visual comparison of the results is portrayed in Fig.~\ref{fig: ablation2}, showing improvement in some of the predicted feature details e.g., near the collar and in the adjacent region of upper and lower body clothes we see improved details.

\begin{table}[ht]
	\centering
	\caption{On the significance of MPN and MPN without the parsing branch in the synthesis stage of our method.}
	\label{tab: ablation2}
	\begin{tabular}{lc} %
		\toprule
		Method & FID$\downarrow$\\
		\midrule
		Ours (target warping w/o mask from MPN) & 16.69\\
		Ours (w/o parsing branch in MPN) & 16.66\\
		Ours & 16.38 \\
		\bottomrule
	\end{tabular}%
\end{table}

\section{Limitations}
\label{limitations}
Our method has a few limitations. First, the quality of inpainting predicted by the proposed ISN is not satisfactory in some cases as shown in Fig.~\ref{fig: limitations}(a). Second, as observed in the quantitative analysis presented in Sec.~\ref{experiments}, our performance on the VITONHD~\cite{vitonhd} dataset is lower compared to that on the MPV dataset. We see one of the main reasons behind this is the poor fit of the target clothing which is predicted by the proposed MPN. See Fig.~\ref{fig: limitations}(b) where the target clothing is full-sleeve but the predicted target mask has half-sleeve. We train MPN on same-person multi-view images~\footnote{Details given in supplementary materials}. We could not train MPN on VITONHD as it does not contain such data. Thus the inaccurate target clothing mask predicted by MPN might be due to the limited performance of MPN across datasets. In future, we aim to delve deeper into these issues.

\section{Conclusion}
\label{sec: conclusion}
This work proposes a novel model-to-person virtual try-on approach that attempts to propose a solution that is robust in terms of pose variation. The challenge lies majorly in handling complex human postures. Most of the previous methods employed one transformation function to compute the target transformation of the source clothing. But different clothing parts should be warped independently to model varied natures of movement of different parts of the body, and better handle the cases of overlap among different parts. To address this, we follow a part-specific warping approach. Additionally, we propose a transformation function called AnaTomy-Aware Geometric (ATAG) transform to warp the sleeve that overcomes several issues of previous methods. 
Our warping method is guided by pose key points and follows constraints of human body movements. We also propose two learning-based modules that aid in computing the warp more efficiently and synthesizing a seamless output while also taking care of the occluded regions in the source clothing. While performance improvement is obtained, the computation of landmarks and parsing adds to the overhead. Considering the utility of the proposed geometric features, in the future, we plan to work towards exploring its potential in guiding appearance flow-based methods toward better handling critical arm postures. Additionally, exploring producing natural folds in the target warping will be an exciting direction of work. Also, the part-based warping concept can be used to synthesize new clothing with different attributes taken from different clothing.

\bibliographystyle{IEEEtran}
\bibliography{IEEEabrv,bibliography}

\begin{IEEEbiography}
[{\includegraphics[width=1in,height=1.25in,clip,keepaspectratio]{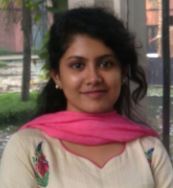}}]{Dr. Debapriya Roy}
is currently working as a Post Doctoral Fellow at TCG – Centres for Research and Education in Science and Technology (TCG-CREST), Kolkata. She received her B.Tech in computer science and engineering from Jalpaiguri Govt. Engg. College, West Bengal, India and M.Tech in information technology from the Indian Institute of Engineering Science and Technology (IIEST), Shibpur, India. She has completed her Ph.D. in Computer Science from the Indian Statistical Institute, Kolkata, India. Her broad areas of research include computer vision, image processing, and deep learning.
\end{IEEEbiography}

\begin{IEEEbiography}[{\includegraphics[width=1in,height=1.25in,clip,keepaspectratio]{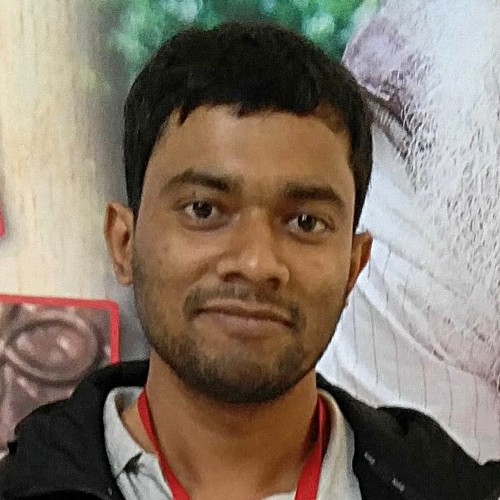}}]{Dr. Sanchayan Santra} 
is a researcher at the Institute for Datability Science, Osaka University. He has completed his Ph.D. in computer science from Indian Statistical Institute, Kolkata, India. Before that, he was at Ramakrishna Mission Vivekananda Educational and Research Institute, Belur Math, West Bengal, India, pursuing his master’s degree. His research interests include Computational Photography, Computer Vision, and Image Processing.
\end{IEEEbiography}

\begin{IEEEbiography}[{\includegraphics[width=1in,height=1.25in, clip,keepaspectratio]{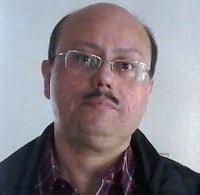}}]{Prof. Diganta Mukherjee} did his BStat, MStat, PhD (Economics) all from Indian Statistical Institute. His research interests are Welfare \& Development Economics in general and multidimensional poverty in particular. He is also active in the area of Finance, particularly mathematical models for pricing contracts and analysis and modeling of Network data. He is an ex-faculty of Jawaharlal Nehru University, Essex University, and ICFAI Business School, and is now a faculty at the Indian Statistical Institute, Kolkata, India. He has 100 research publications in national and international journals and authored four books. He has been involved in projects with large corporate houses and various ministries of the Government of India and West Bengal Government and also consults for NGOs and other Public Action Groups. He is presently, acting as a technical advisor to Multi-Commodities Exchange, Reserve Bank of India, Securities and Exchange Board of India, National Sample Survey Office, National Accounts Division (Central Statistical Office), UNDP (through Ministry of Home Affairs), Ministry of Electronics \& Information Technology, Ministry of Health and Family Welfare.

\end{IEEEbiography}

\begin{IEEEbiography}[{\includegraphics[width=1in,height=1.25in,clip,keepaspectratio]{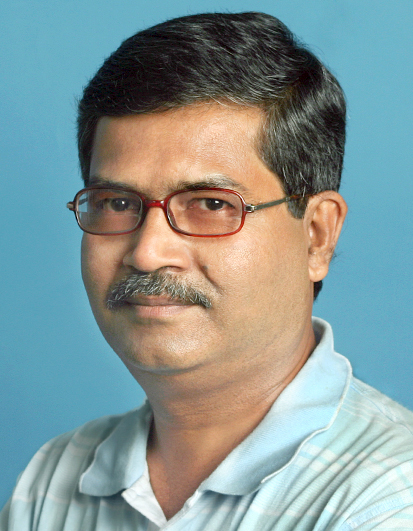}}]{Prof. Bhabatosh Chanda} received the
B.E. degree in electronics and telecommunication
engineering and the Ph.D. degree in electrical engineering from the University of Calcutta, Kolkata,
India, in 1979 and 1988, respectively. He is currently a Professor with the Indian Institute of Information Technology Kalyani and formerly
with the Indian Statistical Institute,
Kolkata, India. He received the Young Scientist
Medal of the Indian National Science Academy in
1989, the Vikram Sarabhai Research Award in 2002,
and the IETE-Ram Lal Wadhwa Gold Medal in
2007. He is a fellow of the Institute of Electronics
and Telecommunication Engineers, the National Academy of Science, India,
the Indian National Academy of Engineering, and the International Association of Pattern Recognition.
\end{IEEEbiography}

\end{document}